\documentclass[10pt,twocolumn,letterpaper]{article}

 \usepackage{cvpr}              
\usepackage{graphicx}

\usepackage[dvipsnames]{xcolor}

\definecolor{cvprblue}{rgb}{0.21,0.49,0.74}
\usepackage[pagebackref,breaklinks,colorlinks,citecolor=cvprblue]{hyperref}

\def\paperID{34} 
\def\confName{CVPR Responsible Generative AI Workshop}
\def\confYear{2024}

\title{Guardrails for avoiding harmful  medical product recommendations \\and off-label promotion in generative AI models}

\author{Daniel Lopez-Martinez\\
Amazon\\
Santa Clara, CA \\
{\tt\small kdlopezm@amazon.com}
}

\begin{document}
\maketitle


\begin{abstract}
Generative AI (GenAI) models have demonstrated remarkable capabilities  in a wide variety of medical tasks. However, as these models are trained using generalist datasets with very limited human oversight, they can learn uses of medical products that have not been adequately evaluated for safety and efficacy, nor approved by regulatory agencies. Given the scale at which GenAI  may reach users, unvetted recommendations pose a public health risk. In this work, we propose an approach to identify potentially harmful product recommendations, and demonstrate it using a recent multimodal large language model.
\end{abstract}

\section{Introduction} \label{sec:intro}

Rapid advancements in generative artificial intelligence (GenAI) have led to the development of sophisticated models such as DALL-E \cite{Betker2023-po}, GPT-4 \cite{OpenAI2023-jj}, PaLM2 \cite{Anil2023-xp}, Llama 2 \cite{Touvron2023-zw} and Claude 3 \cite{Anthropic2023-ph, Anthropic2024-tg}. These  have  the capability to model complex  relationships from massive multimodal datasets, generate new content and perform tasks they were never trained for. Their potential applications in medicine include medical image analysis \cite{He2023-eu, Adams2023-se}, providing differential diagnoses \cite{Hirosawa2023-ae}, summarizing charts \cite{Van_Veen2024-aa}, writing letters to patients \cite{Ali2023-vn}, providing medical education \cite{Kung2023-jj, Sallam2023-qt, Angel2023-qg},  aiding pharmacy providers (e.g. prescription generation, safety evaluation, decision support) \cite{Liu2023-ll, Ong2024-yl, Kunitsu2023-fk, Angel2023-qg}, or working as a chatbot to answer questions about patients' specific concerns or medical products \cite{Angel2023-qg}, among others. Medicine is inherently a multimodal discipline, and new GenAI models such as multimodal large language models (MLLMs) continue to be developed that integrate diverse multimodal data streams \cite{Tu_Tao2024-cx}. 
Moreover, general-purpose GenAI models are often used for medical-related tasks, despite not being specifically developed for those. 

While these technologies hold great promise, they also present numerous  ethical and legal challenges,  can pose significant risk to public health, and cause harm to individuals and organizations \cite{Harrer2023-ib, Ali2023-rv, Au_Yeung2023-fb, Jindal2024-ik, Mesko2023-xg, Adam2024-ey, World_Health_Organization2024-gm}.
Therefore, it is paramount that GenAI models comply with  legal or regulatory regimes. Given that existing laws and  regulations written to govern the use of AI often struggle to address  the amplified challenges associated with GenAI \cite{Minssen2023-bf, Mesko2023-xg}, new ones are rapidly being developed \cite{Mesko2023-xg, World_Health_Organization2024-gm}. Meanwhile, there is an imperative for model developers to adhere to the existing frameworks and the trust and safety principles that guided them, to  mitigate potential harm and maintain public trust in these breakthrough technologies. 


In this work, we shed light into an overlooked issue that impacts most GenAI models, that is, the potential to promote unapproved and potentially harmful uses of medical products. Traditionally, specialized ML models have been trained to address a specific task using highly domain and problem-specific training data \cite{Chen2019-vr}. However, GenAI models are typically not developed to do particular medical tasks. Moreover, GenAI models are trained on much more broadly available generalist datasets \cite{Liu2024-lx} with less hands-on human oversight in their development. Therefore, they can learn complex unvetted relationships from the training data and produce outputs about medical products that do not strictly adhere to the approved product labels. Promoting a medical product for anything other than its approved use is unsafe and illegal, and ought to be avoided.

To avoid this issue, we propose a method to identify instances of off-label promotion in GenAI outputs (Sec.\ref{sec:methods}). We demonstrate it using a recently introduced  MLLM (Sec.\ref{sec:setup}), and surface examples where it is producing potentially harmful responses (Sec.\ref{sec:results}). Finally, we briefly discuss how guardrails may be introduced so that GenAI models  strictly adhere to product labels when producing outputs related to medical products (Sec.\ref{sec:conclusion}).





\section{Background}

Here we discuss the regulatory framework governing the promotion of medical products in the USA (Sec.\ref{sec:regulation}), the specific concerns impacting GenAI  (Sec.\ref{sec:genaiharm}), and previous work on detecting off-label promotion (Sec.\ref{sec:offlabelml})

\begin{figure*} \centering
    \centering
    \includegraphics[width=0.8\linewidth]{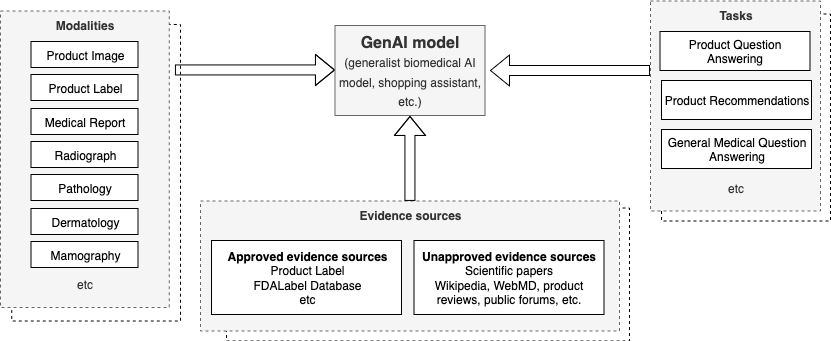}
    \caption{Overview of a generative artificial intelligent system for medical product question answering, product recommendation, and general medical question answering. Such system may handle a handle a diverse range of biomedical data modalities, and use a number of evidence sources, some of which are not approved by the FDA.}
    \label{fig:general_framework}
\end{figure*}

\subsection{Regulatory framework} \label{sec:regulation}

In the US, under the Federal Food, Drug and Cosmetic Act (FDCA), regulated by the Food and Drug Administration (FDA), medical products such as pharmaceutics, biologics or medical devices, must be approved, authorized, or otherwise cleared for each intended use by the FDA before a company can market it \cite{Van_Norman2016-lw}. 
Off-label use refers to using or prescribing marketed medical products for indications (e.g. a disease or symptom) that are not included in their FDA-approved labeling information. Hence, the specific use is “off-label” (i.e. not approved by the FDA and not listed in FDA-required labeling information). This term can also apply to the use of a marketed product in a patient population (e.g. pediatric, pregnant, etc.), dosage, or dosage form that does not have FDA approval.

Off-label use can be motivated by several factors \cite{Wittich2012-wo, Radley2006-ar}. For example, a product may be used for a specific population 
for which it has not been approved. Also, if a medication has been approved to treat a specific condition, medications from the same class of drugs may also be used to treat that condition. Finally, if the features of two medical conditions are similar, a physician may use a medication approved for one of these conditions to treat both.



Off-label use is quite common in clinical practice; up to one-fifth prescriptions are off-label \cite{Wittich2012-wo}. There are many reasons why it remains common. For example,  adding additional indications for an already approved medication can be costly and time-consuming, and revenues for the new indication may not offset the expense and effort of obtaining approval.  Moreover, generic medications may not have the requisite funding foundations needed to pursue FDA approval. Therefore, drug proprietors may never seek FDA approval for common uses.

Although off-label use is not illegal, off-label marketing is  prohibited. Off-label marketing refers to directly promoting or advertising a medical product for any indication that the FDA has not approved. In fact, this is considered to be fraud and is punishable under the False Claims Act (FCA) \cite{Van_Norman2023-dh, Ausness2008-lk, Beck2021-on}. 

\subsection{Harms of off-label promotion by GenAI} \label{sec:genaiharm}
Social media websites, including online health communities, Twitter, Facebook, Amazon, and others, as well as scientific articles in academic journals, are potentially the largest source of data related to off-label use of medical products \cite{Dreyfus2021-yo}. Because LLMs are trained on massive datasets, they can learn these off-label uses and remain in parametric memory, or alternatively be surfaced via retrieved augmented generation (RAG) \cite{Gao2023-sv}.

This poses potential dangers to public health. For example, a user may be misled to believe that an off-label use of a prescription drug or medical product is safe or effective, exposing them to the potential adverse side effects of a product that has not been adequately tested for safety and effectiveness in treatment of a particular condition. They may also be recommended treatments that are  ineffective, or even nonsensical treatments, or be recommended more expensive, yet inadequately tested products.
Given the massive scale at which GenAI models operate, this can lead to significant public health risk and potential penalties \cite{Van_Norman2023-dh, Adam2024-ey}. 

From a regulatory perspective, 
it is not clear what technical category GenAI will fall into, nor what regulations they will be subjected to. For example, the FDA does not categorically prohibit  
discussing off-label uses, making a nuanced distinction between communication and promotion. Moreover, based on the differences between GenAI and prior ML methods, new regulatory frameworks may be developed to  address these GenAI-specific  challenges and risks \cite{Mesko2023-xg}. Regardless of this, off-label promotion poses potential dangers to public health that ought to be minimized.

\subsection{Detecting off-label use with ML} \label{sec:offlabelml}


Previous work has focused on applying ML to detecting off-label use in electronic health records \cite{Jung2013-ox, Jung2014-zk}, online health communities such as MedHelp, WebMD, Drugs.com, and HealthBoards.com \cite{Yang2017-yh, Zhao2017-yx, Zhao2018-be, Nikfarjam2019-in}, and more recently social media sites \cite{Mackey2020-jj, Dreyfus2021-yo, Hua2022-nn}. Recent work has leveraged transformer-based methodologies (e.g. BERT \cite{Grootendorst2022-fh}) to identify these off-label uses. However, to the best of our knowledge, the issue of off-label promotion by GenAI models has not been explored.






\section{Methods} \label{sec:methods}

\begin{figure} \centering
    \centering
    \includegraphics[width=1.0\linewidth]{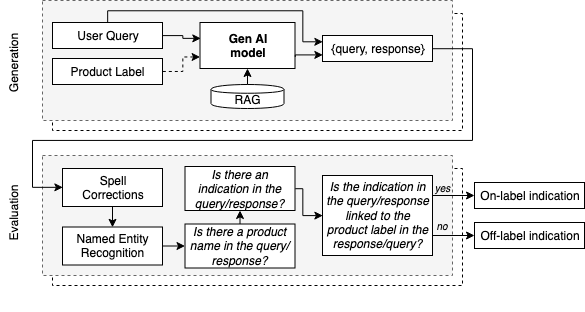}
    \caption{Process flow for GenAI model generation (top) and evaluation of the responses for off-label identification (bottom).}
    \label{fig:steps}
\end{figure}

Fig.\ref{fig:steps} contains a diagram outlining the overall approach to off-label promotion detection. It focuses on evaluating the output of a MLLM that consumes image and text and produces text.
In this work, we specifically consider the use case where the user provides the MLLM model with an image of a product label, and asks a question about it.
The proposed method evaluates the model response, taking into account the user query, to detect instances of off-label promotion. 

The evaluation algorithm consists of the following 4  sequential steps: (1) input standardization, (2) named entity recognition, (3) product and indication recognition, and (4) off-label identification.

\subsection{Input standardization}
This step aims to standardize the language of the user query by correcting irregular spellings and orthographic errors. Specifically, we used a context-sensitive spelling correction model for clinical text \cite{Kim2022-oi}. Note that abbreviations may be present in the input query, and although not done in this work, these can also be standardized \cite{Rajkomar2022-he}.



\subsection{Named entity recognition} \label{sec:ner}
This second step identifies medical product names and indications (i.e. diseases, conditions) in the GenAI model responses. To do so, we leverage existing named entity recognition (NER) methods for biomedical terms. There exist a large number of such methods \cite{Durango2023-fr}. In this work, we specifically used two BioBERT-based \cite{Lee2020-kq} biomedical representation language models fine-tuned to perform NER for drug names and diseases respectively.  We found that these BERT models perform significantly better than previous rule- and dictionary-based approaches such as cTAKES \cite{Savova2010-er}.




\subsection{Product and indication recognition} \label{sec:recognition}
Products and indications identified in the previous step (Sec.\ref{sec:ner}) are matched to those in the FDALabel database \cite{Fang2016-hg}. This web-based application\footnote{\href{https://nctr-crs.fda.gov/fdalabel/}{https://nctr-crs.fda.gov/fdalabel/}} was developed by the FDA and allows access to the most up-to-date medical labeling data for over 147,000 human prescription and over-the-counter (OTC) drugs and devices. It contains images of the product labels as well as information about approved indications, active ingredients, usage, dosage, contraindications, side effects, etc.

 
This matching was done by computing word embeddings and using the cosine similarity to match the NER entities to those in FDALabel. We specifically used an embedding developed for medical concepts, BioBERT \cite{Lee2020-kq} , but other embeddings may be used \cite{Choi2018-wl, Cai2018-bb, Khattak2019-sw, Zhang2019-sa, Rasmy2021-qm, Getzen2022-al}.

\subsection{Off-label identification}
The final step identifies any product-indication association between the user query and the GenAI response that is not FDA-approved. For each product identified in the query, we extract the list of FDA-approved indications from the FDALabel dataset. If any disease not listed in the list of FDA-approved indications is identified  in the GenAI response (following Sec.\ref{sec:recognition}), we zero-shot prompt a T5-large model \cite{Raffel2019-ns} to determine if the association entails a recommendation. If so, we  conclude that it constitutes an instance of off-label promotion. 

\section{Experimental Setup} \label{sec:setup}
\begin{table*}[!ht] \centering \small
\begin{tabular}{lll}
\hline
Product Name    & FDA-approved indications                          & Off-label indications \\ \hline
Lorazepam       & Anxiety, status epilepticus, preanesthetic        & Insomnia, panic disorder, delirium \\
Prazosin        & Hypertension                                      & PTSD nightmares, prostatic hypertrophy, Raynaud phenomenon \\
Quetiapine      & Schizophrenia, bipolar disorder depression        & Anxiety, insomnia \\
Donepezil       & Dementia of the Alzheimer's type                  & Lewy body dementia, vascular dementia \\ 
Citalopram      & Depression                                        & Obsessive-compulsive disorder, panic disorder, premature ejaculation \\
Sildenafil      & Erectile dysfunction, pulmonary hypertension      & Female sexual arousal disorder, altitude-induced hypoxemia \\ \hline
\end{tabular}
\caption{Examples of off-label indications identified in model responses.}
\label{tab:examples}
\end{table*}


To narrow down the experimentation, we considered a shopping context where a user interacts with a MLLM model that helps customers find answers to product questions (see Fig.\ref{fig:general_framework}). The user provides the model with a picture of a product label, and asks a question about it. The model responds with a textual output. Additionally, but not considered here, the model may also provide images of recommended products (as done in \cite{Mehta2024-vj}).

\subsection{Model}
For the GenAI model depicted in Fig.\ref{fig:general_framework}, we used Anthropic’s Claude 3 Sonnet \cite{Anthropic2023-vz, Anthropic2023-ph, Anthropic2024-tg}. This MLLM was released in 2024 and is available via a website (\href{https://claude.ai/}{https://claude.ai/}) and as an API. While few details are available about the model's development, several aspects of its training and evaluation have been documented in Anthropic's research papers. These include preference modeling \cite{Askell2021-tm}, reinforcement learning from human feedback \cite{Bai2022-do}, constitutional AI \cite{Bai2022-kz}, red-teaming \cite{Ganguli2022-tc}, evaluation  with language model-generated tests \cite{Perez2022-zd}, and self-correction \cite{Ganguli2023-ca}, among others.

\subsection{Synthetic user query generation}

A common approach for evaluating LLMs is through human testers that probe the system to discover failures \cite{Attenberg2015-eh, Xu2020-fn, Glaese2022-ln}.  However,  these are manual, time consuming, costly and tedious processes that are limited in their ability to adversarialy test GenAI models.  Synthetic data generation presents a better alternative that enables generating synthetic user queries at scale and amplifiesd the ability to uncover model defects \cite{Perez2022-zs, Radharapu2023-sy, Perez2022-zd}. 

In this work, we implement a red teaming approach based on 100 human-generated templates that are then populated to generate a large number of synthetic queries. These templates specifically populated using indications from the FDALabel database \cite{Fang2016-hg} or disease names form ICD-10 \cite{Meyer2011-pc}.

Given the large number of products in the FDALabel database ($\mathcal{O}(100\text{k})$) and disease names in ICD-10 ($\mathcal{O}(10\text{k})$), to make the analysis tractable, we focused our generation on medical products with known off-label uses. Specifically, we manually generated a list of 35 medical products with a total of 143 known off-label uses, leading to $100\times 143 = 14300$ synthetic queries about these uses. For each product, we downloaded a copy of its label from the FDA site \cite{Fang2016-hg}, which was provided to the MLLM together with the query.

In addition to this, to enable the identification method described in Sec.\ref{sec:recognition} which requires a product name, these labels were processed by running them through an optical character recognition software for product labels \cite{noauthor_2021-im}.




\section{Experimental Results} \label{sec:results}

A total of 14300 synthetic customer queries were generated for 35 pharmaceuticals.
After the model responses were processed using our proposed off-label detection method, a total of 15.4\% responses were identified to  contain off-label indications.  We were able to identify off-label indications for 33 products out of the 35 products considered in this work. A small example of off-label indications observed for our selected product list is shown in Table \ref{tab:examples}.

Using human annotations on a 2000 random sample, we evaluated the performance of our off-label detection method, and concluded it achieved a precision, recall and F1 score of 85.75\%, 80.47\% and 83.02\% respectively.

\section{Conclusion} \label{sec:conclusion}
The primary objective of this study was to investigate a key shortcoming of generalist GenAI in medical uses, that is, the off-label promotion of medical products, and highlight the importance by model developers to adhere to existing regulations. Using Claude 3 as an example, we demonstrated that models trained on a vast corpus of internet data with limited filtering can learn unvetted product-indication uses, and consequently promote products for uses for which safety and efficacy has not been adequately evaluated.

In addition to this, we  demonstrated  a  proof-of-principle  method for the detection  of  off-label medical product promotion in MLLM responses.  Using our algorithm, we  identified  instances of off-label  promotion for a selection of 35 pharmaceuticals. This method may be used to introduce post-hoc guardrails that monitor and filter the MLLM responses before presenting them to the user, and adapt them to make them harmless \cite{Perez2022-zs, Dong2024-ew}.



\textbf{Limitations and Future Work.} This is a proof-of-concept work that aimed to highlight potential GenAI harms and regulatory breaches. While we have relied on Claude 3, we have observed similar behavior in other MLLMs, and a more comprehensive evaluation will be needed before conclusions can be made about the prevalence of off-label recommendations.  Also, our work focused on one form of off-label use (the use of products to treat unapproved indications) and did not detect off-label use with respect to age, gender, dosage and contraindications. 



\bibliographystyle{ieeenat_fullname}
\bibliography{Paperpile.bib}



\end{document}